\documentclass{article}
\usepackage{spconf,amsmath,amssymb,graphicx}
\usepackage{booktabs,multirow}
\usepackage[hidelinks]{hyperref}
\usepackage{microtype}

\graphicspath{{figures/}}
\newcommand{\IC}{\mathrm{IC}}

\title{INFORMATION CAPACITY OF GENERATIVE VIDEO COMPRESSION:\\
QUANTIFYING THE RATE--COMPUTE EXCHANGE AT IDENTICAL QUALITY}

\name{Cheng Yuan, Jiawei Shao, Xuelong Li\thanks{Corresponding author: Xuelong Li (xuelong\_li@ieee.org).}}
\address{Institute of Artificial Intelligence, China Telecom (TeleAI)}

\begin{document}
\ninept
\maketitle

\begin{abstract}
Under the AI Flow framework, communication networks distribute intelligence across devices, edge servers, and cloud clusters, and computation at the receiver becomes a resource that can substitute for transmitted bits. Generative video compression (GVC) embodies this exchange by sending compact tokens with ultra-low bitrate and letting a generative decoder synthesize the video. To quantify such exchange, we define information capacity (IC) as the negative logarithmic slope along an iso-quality contour, namely the fraction of rate saved per fractional increase in compute at identical quality. IC is dimensionless and unit-invariant, thus enabling an architecture-agnostic comparison. It forms a field over the operating plane, locating where additional denoising steps are worth their cost. For two GVC decoders with adjustable test-time compute, a rate--compute power law fits DISTS with mean error below 3\%, which is used for analytical IC derivations. IC varies across five datasets and two decoders, and the median local IC ratio of a 14B decoder versus a 1.3B decoder is 9.3 at the systems' respective grid centers.
\end{abstract}


\section{Introduction}
\label{sec:intro}

AI Flow envisions communication networks that distribute intelligence rather than bits, coordinating end devices, edge servers, and cloud clusters so that inference and transmission are optimized jointly~\cite{an2026aiflow,shao2026edge}. A recurring principle of this framework is that computation can be traded for bandwidth. A receiver equipped with a strong generative prior requires fewer transmitted bits to reproduce the content, as the transmitted bits need only steer the prior rather than describe every pixel. Generative video compression (GVC) is a direct realization of this principle~\cite{chen2026gvc}. The encoder transmits compressed keyframes and compact latent tokens, and a diffusion transformer (DiT) at the receiver synthesizes the video, reaching compression rates as low as 0.02\%. At such rates conventional codecs such as HEVC~\cite{sullivan2012hevc} collapse into blocking and smearing, whereas GVC preserves sharp textures and coherent motion, so that its perceptual quality at a given bitrate exceeds that of HEVC at several times the bitrate~\cite{chen2026gvc}. 

Recent diffusion-based codecs report the same advantage. Video-native diffusion priors remove the flickering of frame-wise generative codecs below 0.01 bits per pixel (bpp)~\cite{mao2026gnvc}, single-step diffusion decoders cut sampling cost while retaining the perceptual gain~\cite{xue2026s2vc}, and lightweight semantic side information enables reconstruction at roughly 0.003 bpp~\cite{zhang2026diffusion}. Perceptual quality in the ultra-low-bitrate regime is thus increasingly determined by what the decoder can compute rather than by what the channel can carry.

Lagrangian rate--distortion--complexity (RDC) optimization controls motion-search effort in complexity-scalable encoding~\cite{wu2007rdc} and accounts for decoding complexity or energy in mode decisions~\cite{mallikarachchi2020complexity,herglotz2019derdo}. Standard codec comparison uses BD-rate~\cite{bjontegaard2001}, which measures rate savings between two codecs at equal quality but ignores the differences in decoder complexity. Scaling laws in language modeling relate loss to training compute~\cite{kaplan2020scaling,hoffmann2022chinchilla}, and studies of diffusion samplers characterize how sample quality improves with the number of denoising steps~\cite{song2021ddim,karras2022edm,salimans2022progressive}. For a large language model (LLM), the information capacity metric~\cite{yuan2025ic} relates text compression performance to computational complexity. To our knowledge, the local bitrate savings at fixed perceptual quality have not been quantified for generative decoders with an adjustable test-time denoising budget. This quantity bears on how many denoising steps the receiver should spend under a given uplink budget.

This paper addresses this gap. We treat a GVC method as a two-input system whose reconstruction quality $Q$ depends on rate $R$ and decoder compute $C$, and define the information capacity $\IC(R,C)$ as the negative logarithmic slope of the iso-quality contour in the $(R,C)$ plane. Our contributions are summarized as follows.
\begin{itemize}\setlength{\itemsep}{0pt}
  \item We introduce one possible definition of information capacity $\IC(R,C)$ for a generative codec, namely the elasticity of rate with respect to decoder compute at constant quality. It is dimensionless and invariant to the units of $R$ and $C$, and can be obtained analytically from a fitted surface or empirically from interpolated iso-quality contours. It changes with $R$ and $C$, constituting a field over the operating plane, which exposes where extra compute is worth its cost.
  \item We measure the DISTS quality surface for the 1.3B and 14B GVC decoders~\cite{chen2026gvc} with 760 configurations of keyframe quantization parameter (QP) and denoising steps on five datasets, and show that a two-factor power law with a distortion floor fits measured DISTS with a mean error below 3\%.
  \item We derive $\IC$ over the rate--compute plane and show that the fitted local $\IC$ ratio of a 14B decoder and a 1.3B decoder has a median of 9.3 at the respective grid centers, and that $\IC$ varies across datasets, indicating imbalanced performance on the rate--compute trade-off in GVC methods.
\end{itemize}

\begin{figure*}[t]
  \centering
  \includegraphics[trim=0 6bp 0 5bp,clip,width=\textwidth]{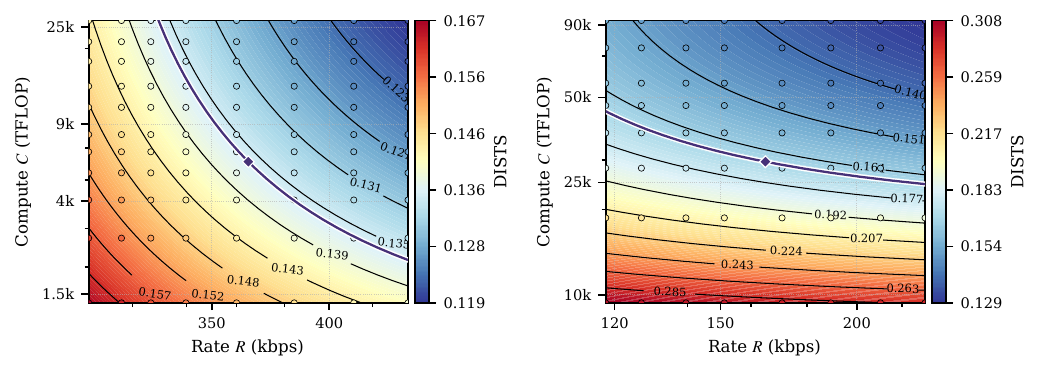}
  \caption{Fitted DISTS surface from Eq.~\eqref{eq:surface} on MCL-JCV for GVC-1.3B (left) and GVC-14B (right) on logarithmic rate--compute axes with iso-quality contours. Circles show measured DISTS. The purple line marks the in-grid contour through the center (diamond). Contours of the 1.3B decoder are steep, so quality is governed mainly by rate, whereas those of the 14B decoder are nearly horizontal at few denoising steps and bend toward the rate axis only at higher steps.}
  \label{fig:surface}
\end{figure*}

\section{Methods}
\label{sec:method}

\subsection{Rate--compute--quality surface}
\label{sec:surface}
A GVC configuration is specified by the bits it transmits, the operations its decoder performs, and the reconstruction quality. We denote by $R$ the bitrate of the complete bitstream in kbps, by $C$ the decoder compute required to reconstruct a sequence in TFLOP, and by $Q$ a distortion metric for which lower is better. Adjusting the encoder-side rate control sweeps over $R$, and changing the decoder-side compute control sweeps over $C$. Each configuration yields one measured triple $(R,C,Q)$ after averaging over the sequences of a dataset.

This formulation is deliberately agnostic to the internal processing details of a codec. The analysis below applies unchanged to any GVC implementation that exposes configurable hyperparameters during inference, with respect to both data rate (e.g., keyframe QP, latent quantization, token budget) and computational complexity (e.g., denoising steps, decoder size, sampler order). In this paper the rate-related hyperparameter is the keyframe QP, and the compute is determined by the number of denoising steps of the DiT decoder. The surface and the metric derived from these results can be generalized to other GVC designs with different configurable hyperparameters.

Following the functional form that describes compute--loss scaling in large models~\cite{kaplan2020scaling,hoffmann2022chinchilla}, we model the distortion as a two-factor power law with a floor,
\begin{equation}
  Q(R,C) = a\,\Big(\tfrac{R}{R_0}\Big)^{-b} + c\,\Big(\tfrac{C}{C_0}\Big)^{-d} + e,
  \label{eq:surface}
\end{equation}
with $a,c\ge 0$, $b,d>0$, and $e\ge 0$. The first term captures the distortion removed by transmitting more bits, the second the distortion removed by spending more decoder compute, and $e$ the fitted asymptotic distortion for the encoder representation and decoder prior. The additive form encodes the assumption that the two resources remove largely distinct components of distortion, each with diminishing returns. Sec.~\ref{sec:results} reports the resulting fit errors and checks the contour slopes against measurements.

\subsection{Information capacity}
\label{sec:ic}
An iso-quality contour of the surface is the set of $(R,C)$ that achieves identical quality, expressed as $Q(R,C)=Q^\star$. We define the information capacity of the codec at operating point $(R,C)$ as the negative logarithmic slope of the contour through that point:
\begin{equation}
  \IC(R,C) \triangleq -\left.\frac{d \ln R}{d \ln C}\right|_{Q = Q(R,C)},
  \label{eq:ic}
\end{equation}
in analogy with the information capacity of an LLM, which likewise relates compression performance to computational complexity~\cite{yuan2025ic}. 

$\IC$ can be evaluated in two ways. Empirically, one interpolates $\ln R$ between neighboring QPs at DISTS $Q^\star$ for each of two step counts and computes $-\Delta\ln R/\Delta\ln C$ across the resulting points. Analytically, one sets $\mathrm{d}Q=0$ in~\eqref{eq:surface}, and implicit differentiation yields the closed form
\begin{equation}
  \IC(R,C) = \frac{c\,d\,(C/C_0)^{-d}}{a\,b\,(R/R_0)^{-b}} ,
  \label{eq:ic_fit}
\end{equation}
which gives the local elasticity on the fitted surface.

Locally, one percent more compute saves approximately $\IC$ percent in rate. For $k\ge1$, solving $Q(R_k,kC)=Q(R,C)$ in~\eqref{eq:surface} with $I=\IC(R,C)$ gives
\begin{equation}
  \eta_k=1-\frac{R_k}{R}=1-\left[1+\frac{bI}{d}\left(1-k^{-d}\right)\right]^{-1/b}.
  \label{eq:saving}
\end{equation}
Along this contour, $\IC$ decreases as compute rises and rate falls. Using the starting $\IC$ throughout would therefore overstate the fitted saving $\eta_2$ for doubling decoder compute.

Three properties of IC can be observed from~\eqref{eq:ic_fit}. First, $\IC$ is dimensionless and invariant to rescaling either $R$ or $C$, so GVC methods with different operating rate--compute ranges or group-of-pictures (GOP) lengths remain comparable. Second, $\IC\propto R^{\,b}C^{-d}$, so compute is most valuable at high rate, where the rate term has already saturated, and at low compute, where the compute term still dominates. The change rate is governed by the exponents $b$ and $d$ in general. Third, $\IC$ is a field rather than a number. Its value changes with both $R$ and $C$, hence the same codec can offer a favorable exchange of compute for bits at one operating point and a negligible one at another. Any scalar quoted for a codec must therefore be tied to a stated operating point.

\begin{figure*}[t]
  \centering
  \includegraphics[trim=0 6bp 0 5bp,clip,width=\textwidth]{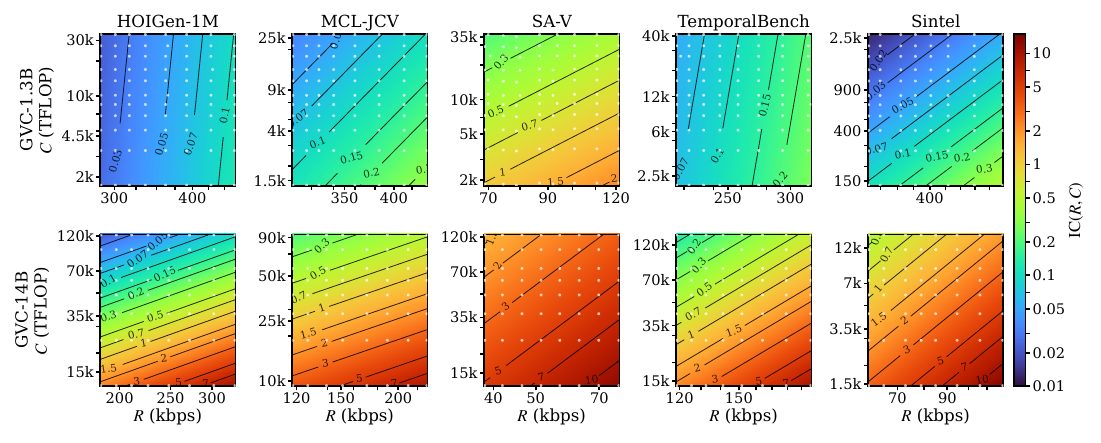}
  \caption{Information capacity field $\IC(R,C)$ from Eq.~\eqref{eq:ic_fit} for GVC-1.3B (top) and GVC-14B (bottom) on five datasets with a shared logarithmic color scale. Black curves are analytical contours of $\IC$ and white dots denote the measured configurations. $\IC$ increases with rate and decreases with compute on every panel. Each panel spans its own measured rate and compute ranges.}
  \label{fig:icmaps}
\end{figure*}

\section{Results}
\label{sec:results}

\subsection{Experimental setup}
\label{sec:setup}

\subsubsection{Models} 
We evaluate the GVC-1.3B and GVC-14B systems from~\cite{chen2026gvc}. Both transmit learned-codec keyframes together with quantized spatio-temporal latents per GOP (the 1.3B system additionally sends a sparse canny edge map) and reconstruct each GOP with a Wan2.1-based DiT decoder~\cite{wan2025} conditioned on the decoded keyframes. The decoders differ in parameter count (1.3B vs.\ 14B) and GOP length (29 vs.\ 45 frames). We refer to~\cite{chen2026gvc} for architectural details and use the inference pipelines of that work without modification.

\subsubsection{Empirical measurements} 
Rate is controlled by the keyframe QP, $\mathrm{QP}\in\{30,34,\dots,58\}$, which changes $R$ by a factor of 1.2--2 across the range depending on the dataset. Compute is controlled by the number of denoising steps $N$, with $N\in\{1,2,3,4,5,6,8,10,13,16,20\}$ for GVC-1.3B and $N\in\{1,\dots,6,8,10\}$ for GVC-14B, constituting 88 and 64 configurations per dataset, respectively. $C$ is the theoretical DiT cost in TFLOP, dominated by attention and MLP FLOPs per denoising pass times $N$ times the number of GOPs in a sequence. Auxiliary components such as VAE and keyframe codec are excluded because they are very small relative to the DiT and independent of $N$.

\subsubsection{Fitting procedure}
The five parameters in Eq.~\eqref{eq:surface} are estimated by nonlinear least squares with relative-error weighting. The normalizers $R_0$ and $C_0$ are the geometric centers of the measured rate and compute ranges, so that $a$ and $c$ are directly the rate- and compute-limited contributions at the center of the grid. Exponents are bounded to $[0.01,20]$, the fit is started from a $7\times7$ grid of initial $(b,d)$ pairs between $0.1$ and $8$, and the solution with the smallest weighted residual is kept.

\subsubsection{Metric and dataset} 
Quality $Q$ is quantified by DISTS~\cite{ding2022dists}, a full-reference perceptual metric that combines structure and texture similarity and is commonly used to evaluate generative codecs. We use five datasets evaluated with both models, namely 588 clips from HOIGen-1M~\cite{liu2025hoigen}, the 30 sequences of MCL-JCV~\cite{wang2016mcljcv}, 305 clips from SA-V~\cite{ravi2025sam2}, 400 clips from TemporalBench~\cite{cai2024temporalbench}, and the 23 sequences of Sintel~\cite{butler2012sintel}. Each $(R,C,Q)$ triple is the mean over the clips of a dataset, and one surface~\eqref{eq:surface} is fitted per dataset and model.

\begin{table*}[t]
  \centering
  \caption{DISTS fits, local IC analysis, and rate savings. Grid centers ($R_0$, $C_0$) are in kbps and TFLOP per sequence, respectively. Err is the mean relative DISTS error. The center contour stays in-grid for $1\le k\le K$. Asterisks mark extrapolated $\eta_2$ estimates ($K<2$).}
  \label{tab:fits}
  \small
  \setlength{\aboverulesep}{0.9pt}
  \setlength{\belowrulesep}{0.9pt}
  \setlength{\tabcolsep}{2pt}
  \begin{tabular*}{\textwidth}{@{\extracolsep{\fill}}ll*{13}{r}@{}}
\toprule
 & & \multicolumn{2}{c}{Grid center} & \multicolumn{6}{c}{DISTS fit} & \multicolumn{3}{c}{Local IC} & \multicolumn{2}{c}{Rate savings} \\
\cmidrule(lr){3-4}\cmidrule(lr){5-10}\cmidrule(lr){11-13}\cmidrule(l){14-15}
Dataset & Model & $R_0$ & $C_0$ & $a$ & $b$ & $c$ & $d$ & $e$ & Err (\%) & $\IC_0$ & 95\% CI of $\IC_0$ & Grid range & $\eta_2$ (\%) & $K$ \\
\midrule
\multirow{2}{*}{HOIGen-1M} & 1.3B & 365.5 & 7591 & 0.0155 & 3.13 & 0.0422 & 0.06 & 0.000 & 0.9 & 0.051 & [0.047,0.056] & 0.022--0.12 & 3.2 & 4.440 \\
 & 14B & 247.3 & 38344 & 0.0160 & 2.51 & 0.0092 & 2.02 & 0.052 & 0.7 & 0.462 & [0.403,0.526] & 0.021--9.95 & 13.3 & 3.162 \\
\addlinespace[1pt]
\multirow{2}{*}{MCL-JCV} & 1.3B & 364.9 & 6060 & 0.0210 & 3.06 & 0.0168 & 0.41 & 0.098 & 0.5 & 0.108 & [0.073,0.161] & 0.033--0.35 & 5.7 & 4.425 \\
 & 14B & 164.9 & 29586 & 0.0275 & 1.32 & 0.0379 & 1.26 & 0.103 & 0.7 & 1.322 & [0.867,1.999] & 0.198--8.84 & 36.1$^*$ & 1.509 \\
\addlinespace[1pt]
\multirow{2}{*}{SA-V} & 1.3B & 91.5 & 8139 & 0.0179 & 1.32 & 0.0290 & 0.54 & 0.097 & 0.5 & 0.662 & [0.538,0.821] & 0.205--2.13 & 26.6$^*$ & 1.760 \\
 & 14B & 54.5 & 39008 & 0.0148 & 1.23 & 0.1326 & 0.57 & 0.057 & 2.7 & 4.155 & [3.565,4.821] & 1.391--12.41 & 67.1$^*$ & 1.117 \\
\addlinespace[1pt]
\multirow{2}{*}{TemporalBench} & 1.3B & 261.9 & 9234 & 0.0183 & 3.56 & 0.0874 & 0.09 & 0.000 & 0.5 & 0.120 & [0.108,0.133] & 0.053--0.27 & 6.9 & 4.441 \\
 & 14B & 152.8 & 44219 & 0.0162 & 2.94 & 0.0439 & 1.22 & 0.080 & 1.9 & 1.118 & [1.019,1.222] & 0.131--9.52 & 27.2$^*$ & 1.537 \\
\addlinespace[1pt]
\multirow{2}{*}{Sintel} & 1.3B & 403.8 & 607 & 0.0131 & 6.78 & 0.0090 & 0.73 & 0.120 & 0.4 & 0.074 & [0.043,0.124] & 0.012--0.45 & 3.5 & 4.397 \\
 & 14B & 84.9 & 4636 & 0.0139 & 2.12 & 0.0831 & 0.86 & 0.103 & 2.1 & 2.419 & [1.616,3.423] & 0.439--13.32 & 45.9$^*$ & 1.251 \\
\bottomrule
\end{tabular*}

\end{table*}

\subsection{Fitted surfaces}
Table~\ref{tab:fits} lists the grid centers, all five fitted parameters, and fit errors for the ten dataset--model pairs, and Fig.~\ref{fig:surface} shows the fitted DISTS surface on MCL-JCV. The two-factor power law reproduces the measured DISTS with a mean relative error between 0.4\% and 2.7\% and a maximum error below 7.1\% over 760 configurations. The rate exponent $b$ lies between 1.2 and 3.6 for nine of the ten pairs and reaches 6.8 for GVC-1.3B on Sintel, whose rate range is the narrowest of all pairs with a factor of 1.24. The fitted floor $e$ lies between 0.05 and 0.12 for eight pairs and vanishes for GVC-1.3B on HOIGen-1M and TemporalBench, where the two power-law terms account for the fitted distortion within the measured range.

The fitted compute exponents differ between the two decoders, with $d\in[0.06,0.73]$ for GVC-1.3B and $d\in[0.57,2.02]$ for GVC-14B. On MCL-JCV, additional denoising steps remove little fitted distortion for the smaller decoder, whose iso-quality contours in Fig.~\ref{fig:surface} are steep. In contrast, the larger decoder is strongly compute-limited at few steps, and its contours bend toward the rate axis only once $N$ exceeds roughly 4.

\subsection{Information capacity over the operating plane}
Fig.~\ref{fig:icmaps} illustrates $\IC(R,C)$ over the measured rate--compute grid of each dataset--model pair in each individual panel. Since $\IC\propto R^{\,b}C^{-d}$, the field is monotonic with respect to both variables. It is largest at high rate and low compute and smallest at low rate and high compute. Additionally, it varies significantly for a single dataset--model pair, by a factor between five and several hundred under different bitrates and FLOPs. The field view thus carries information that no scalar can conclude, namely where in the operating plane the exchange of more compute for fewer bits is favorable.

The level set $\IC=1$ separates a region where, locally, one percent of extra compute leads to more than one percent of rate saving from a region where it saves less. For GVC-14B this boundary runs through the interior of the measured grid on four datasets and is exceeded everywhere on the remaining SA-V dataset. In general, the favorable regime is reachable by operating at few denoising steps and a moderate QP. For GVC-1.3B it is reached only on SA-V, and only at the highest rates and fewest steps. On the other four datasets $\IC$ remains below 0.5 over the whole plane.

At adjacent step counts we compared measured and fitted $\IC$ values at the midpoint of their DISTS overlap, requiring unique in-grid endpoints. All 70 empirical elasticities were positive, with 19\% median absolute relative difference from fitted values.

\begin{figure}[t]
  \centering
  \includegraphics[trim=0 7bp 0 7bp,clip,width=\columnwidth]{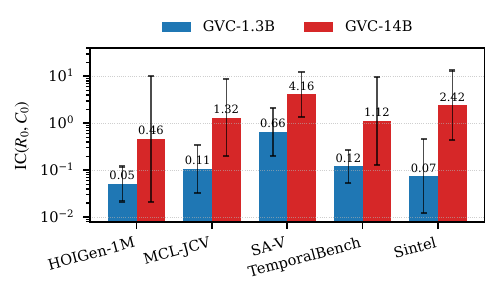}
  \caption{Information capacity at the grid center $\IC(R_0,C_0)$ per dataset and decoder in logarithmic scale. Whiskers indicate the minimum and maximum of $\IC$ over the measured grid.}
  \label{fig:bars}
\end{figure}

\subsection{Cross-model and cross-dataset comparisons}
To compare datasets and decoders, we evaluate the field at each grid center, where~\eqref{eq:ic_fit} reduces to $\IC_0=cd/(ab)$. Fig.~\ref{fig:bars} and Table~\ref{tab:fits} report $\IC_0$ together with its range over the grid. $\IC_0$ ranges from 0.05 to 0.66 for GVC-1.3B and from 0.46 to 4.16 for GVC-14B across five datasets. The 14B versus 1.3B $\IC_0$ ratio ranges from 6.3 (SA-V) to 32.5 (Sintel), with a median of 9.3. Note that the two decoders were measured on different rate ranges, as the 14B bitstreams are 1.5--5 times smaller. Consequently, each $\IC_0$ refers to the operating center of its own model on a specific dataset. Within grids, $\IC$ varies by a factor of 5.1--37 for GVC-1.3B and 8.9--463 for GVC-14B.

We also resample videos in the dataset with replacement while maintaining each video's parameter configurations. Each time we recompute the grid center, refit~\eqref{eq:surface}, and obtain a new $\IC_0$ estimate. The 95\% interval in Table~\ref{tab:fits} spans the 2.5th and 97.5th percentiles of 1,000 estimates per fit. Results show that $\IC_0$ values fluctuate in MCL-JCV and Sintel datasets due to their small sample counts. Refitting without the two outer QPs or step counts changed IC at the original centers by up to 7.2\% or 24.9\%, respectively.

Table~\ref{tab:fits} reports $K=C_{\mathrm{edge}}/C_0$, where $C_{\mathrm{edge}}$ is the compute at which the fitted equal-DISTS contour from $(R_0,C_0)$ first meets the measured grid boundary in the $(R,C)$ plane as compute increases. $K < 2$ holds in five of ten cases, where the doubled compute $2 C_0$ stays in the grid range. For GVC-14B at 0.15278 DISTS on MCL-JCV, bracketed at four and eight steps, interpolation gives 31.2\% savings and the fit gives 34.5\%. Nine of 25 tested 14B doublings had such overlap, with a median absolute savings difference of 2.4 percentage points for measured and fitted values.

The extremes of the dataset ranking coincide for the two decoders, with SA-V having the highest and HOIGen-1M the lowest $\IC_0$ for both decoders. This is largely attributed to the rate exponent $b$. As demonstrated in Table~\ref{tab:fits}, $b\approx1.3$ for both models on SA-V dataset, signifying that additional bits remove distortion slowly and compute is comparatively effective, whereas $b\ge2.5$ on HOIGen-1M, showing that increasing bitrate is effective and the exchange rate for compute is low. This suggests that content on which additional bits are least effective is precisely the content on which decoder compute benefits most, which is the targeted scenario for GVC.

\section{Conclusion}
\label{sec:conclusion}
We have introduced one possible definition of information capacity $\IC(R,C)$ for a generative video codec, namely the elasticity of rate with respect to decoder compute along an iso-quality contour, and derived it in closed form from a two-factor power-law surface. Applied to two GVC decoders on five datasets, the surface fits DISTS to within a few percent. The median local $\IC$ ratio of a 14B decoder and a 1.3B decoder is 9.3 at the systems' respective grid centers. The exchange rate changes across the operating plane and across datasets.

Two limitations bound these conclusions. Compute is measured as theoretical DiT FLOPs, which do not capture memory-bound execution, wall-clock latency, or energy consumption, so the same $\IC$ may correspond to different deployment costs on heterogeneous hardware. Also, $\IC$ is reported on the commonly used metric DISTS alone in this paper. As it is defined relative to a chosen quality metric, exponents fitted to pixel-level fidelity or temporal-consistency metrics may differ.

Future work will extend the analysis to more GVC decoders and different model architectures. It will also validate $\IC$ against downstream-task performance, such that the iso-quality contour directly reflects task utility rather than perceptual distortion. It is also interesting to investigate the adoption of $\IC$ field as a controller for adaptive step allocation across device, edge, and cloud nodes in AI Flow, increasing denoising steps only where $\IC$ exceeds the local price of compute relative to bandwidth.

\bibliographystyle{IEEEbib}
\bibliography{refs}

\section{Compliance with Ethical Standards}
This is a numerical simulation study for which no ethical approval was required.

\end{document}